\documentclass[letterpaper, 10 pt, conference]{ieeeconf}  

\IEEEoverridecommandlockouts                              

\usepackage{newtxmath}
\usepackage{color}
\usepackage{graphicx}
\usepackage{subfig}
\usepackage{float}

\title{\LARGE \bf
Informative Text-Image Alignment for Visual Affordance Learning with Foundation Models
}

\author{Qian Zhang, Lin Zhang, Xing Fang, Mingxin Zhang, Zhiyuan Wei, Ran S, and Wei Zhang$^{*}$
\thanks{*Corresponding author: Wei Zhang}
\thanks{The authors are with School of Control Science and Engineering, Shandong University}%
\thanks{This work was supported by the Key R\&D Program XXX.}%
}

\begin{document}

\maketitle
\thispagestyle{empty}
\pagestyle{empty}

\begin{abstract}

Visual affordance learning is crucial for robots to understand and interact effectively with the physical world. Recent advances in this field attempt to leverage pre-trained knowledge of vision-language foundation models to learn affordance properties with limited training data, providing a novel paradigm for visual affordance learning. However, these methods overlook the significance of maintaining feature alignment between visual images and language descriptions for identifying affordance areas with textual guidance, and thus may lead to suboptimal results. In this paper, we present an informative framework for text-guided affordance learning, which involves information-based constraints to achieve text-image alignment at feature level. Specifically, we design an affordance mutual information  constraint that helps learn appropriate textual prompts and task-oriented visual features simultaneously by maximizing the mutual information between the features of the affordance areas in the input images and the corresponding textual prompts. 
In addition, we propose an object-level information constraint that maximizes the mutual information between the visual features of a given object and the text features of the category it belongs to. This enables the model to capture high-quality representations for the object, providing more reliable semantic priors for identifying affordance regions.
Experimental results on the AGD20K dataset show that the proposed method outperforms existing approaches and achieves the new state-of-the-art in one-shot affordance learning.

\end{abstract}

\section{INTRODUCTION}

Achieving robust and intuitive physical interaction remains a fundamental challenge in robotics. Affordance learning, the process by which robots acquire knowledge about the potential actions objects afford, serves as a cornerstone for enabling machines to perceive their surroundings not just as static scenes, but as opportunities for purposeful interaction, moving closer to human-like competence. Identifying the affordance area of a specific object from images with diverse backgrounds is challenging, primarily because there may be other non-interested objects that cause inference, and the object may support various affordance categories.

Most of the previous works \cite{chuang2018learning, Mur2023bayesian, do2018affordancenet} address the affordance learning task by optimizing neural networks to fit the relationships between image features and affordance labels in a supervised manner, which typically requires large-scale datasets with numerous image-affordance pairs. Although these approaches have achieved promising results, they fail to handle real-world complex scenarios where robots may encounter new categories of object with limited samples. Therefore, enabling models to learn the affordance properties of objects with limited data is crucial for enhancing robots' interaction and manipulation capabilities. To overcome the problem of limited data when facing novel objects, Ning et al. \cite{ning2023whereexplore} investigated the challenging few-shot setting of affordance learning, and build a few-shot affordance learning framework that conducts similarity-guided exploration to handle  unseen objects of novel categories.

In recent years, pre-trained foundation models such as CLIP \cite{radford2021learning} have demonstrated exciting capabilities in representing various kinds of data, including but not limited to images and texts. Leveraging the multimodal priors of foundation models to assist downstream tasks has been an effective and cost-efficient solution, achieving satisfactory task performance with a few labeled data or even a single sample. This offers a new perspective for addressing few-shot affordance learning. Based on this thought, Li et al. \cite{li2024one} proposed one-shot affordance learning (OOAL) with pre-trained models. Specifically, the OOAL framework incorporates the CLIP text encoder, a multi-layer feature fusion module based on DINOv2 \cite{oquab2023dinov2}, and a transformer decoder to generate the final affordance mask. Although this method enhances the transformer decoder by introducing the visual [CLS] token as an extra guidance, it has two drawbacks that may hinder the recognition of text-related affordance regions in images. 
On one hand, it lacks explicit constraints to align the features of textual prompts with those of the visual affordance areas, limiting the ability to identify affordance areas with textual guidance. On the other hand, it overlooks the significant risk that the cross-modal alignment in object level may deteriorate as training progresses,  potentially leading to the visual class token used in the transformer decoder providing negative guidance. 

In order to address the aforementioned drawbacks, we propose a novel affordance learning framework that incorporates mutual information-based constraints to align textual and visual features at both affordance and object levels. At the affordance level, we attempt to align the visual affordance areas with textual descriptions of the corresponding affordance categories by maximizing the mutual information between the features of the affordance regions and those of the affordance-level textual prompts. This serves as the primary basis for locating and identifying the affordance areas with textual guidance.
At the object level, we aim to maintain the text-image alignment properties of pre-trained foundation models, which is achieved by maximizing the mutual information between the visual class tokens and the text features of the corresponding object categories. This constraint ensures that the visual class tokens contain rich semantic information and play a positive role in affordance learning. 
The main contributions of this paper can be summarized in three folds as follows:

\begin{itemize}
    \item We introduce a novel affordance learning framework that employs mutual information-based constraints to achieve text-image alignment between visual affordance areas and textual prompts for affordance categories, thereby facilitating text-guided affordance learning.
    \item We propose object-level mutual information maximization to  preserve the text-image alignment properties of foundation models during training, which provides robust semantic priors for affordance learning.
    \item Comparative experiments and ablation studies demonstrate that the mutual information-based constraints significantly contribute to affordance learning, with our proposed method outperforming existing approaches in affordance learning.
\end{itemize}

\section{Related Works}

\subsection{Visual Affordance Learning}
``Affordance” is originally defined as the opportunity for interaction that an object or the surrounding environment provides. Recently, researchers in the fields of computer vision and robotics have developed numerous approaches to learn affordance properties for various purposes such as robotic manipulation \cite{yang2023watch, li2024learning} and human behavior understanding \cite{bahl2023affordances}. Most of existing methods use CNN-based backbone networks for visual feature extraction, followed by specific heads for affordance classification, detection, and segmentation. These methods rely heavily on large-scale manually annotated dataset, which limits their application in complex scenarios with insufficient labeled data. 

To reduce the reliance on annotated affordance data, some works have explored the weakly supervised and few-shot settings of affordance learning. For example, Cui et al. \cite{cui2023strap} build a structured transformer to achieve affordance segmentation with keypoint supervision.  Ning et al. \cite{ning2023whereexplore} construct a few-shot learning framework to explore affordance properties of novel categories with limited samples, enabling robotic manipulation models generalize well in open-world task scenarios. More recently, some researchers attempt to leverage pre-trained knowledge of foundation models to improve the performance of affordance learning with limited data. For example, 
Li et al. \cite{li2024one} integrate CLIP textual encoder and DINOv2 to build a novel one-shot learning framework, which combines the pre-trained knowledge from the two foundation models to enhance one-shot affordance learning. 

In this paper, we focus on the topic of affordance learning with foundation models and aim to preserve text-image alignment during training with informative constraints to boost the performance of text-guided affordance learning.

\subsection{Foundation Models and Parameter-Efficient Fine-tuning}

In the last several years, foundation models pre-trained with numerous labeled data have stirred significant excitement in the fields of artificial intelligence and robotics. These large-scale foundation models are well-equipped to effectively handle various types of data, including visual, linguistic, and multi-modal inputs. 
In computer vision, the ViT-based pre-trained models, such as DINO and DINOv2 \cite{oquab2023dinov2}, demonstrate powerful capabilities in extracting generalizable visual features. As a typical representative of multi-modal foundation models, CLIP \cite{radford2021learning} and its variants \cite{dong2023maskclip} provide rich visual-text alignment priors, for downstream tasks.

However, adapting these general-purpose models to specialized downstream applications traditionally requires computationally expensive full fine-tuning of all parameters. To overcome this limitation, researchers are focusing on parameter-efficient fine-tuning of foundation models as one of their main areas of interest. Techniques such as prompt tuning \cite{zhu2023prompt}, adapter tuning \cite{gao2024clip}, and low-rank adaptation \cite{hu2022lora} enable fine-tuning pre-trained models with low cost, further enhancing performance on downstream tasks. In the field of affordance learning, Li et al. \cite{li2024one} fine-tuning the CLIP text encoder with learnable prompts to obtain high-quality textual features related to affordance.

In this paper, we leverage the pre-trained knowledge of foundation models and employ prompt tuning techniques with mutual information-based constraints to effectively adapt these models for the affordance learning task.

\begin{figure*}[htbp]
  \centering            
  \includegraphics[width=1.0\textwidth]{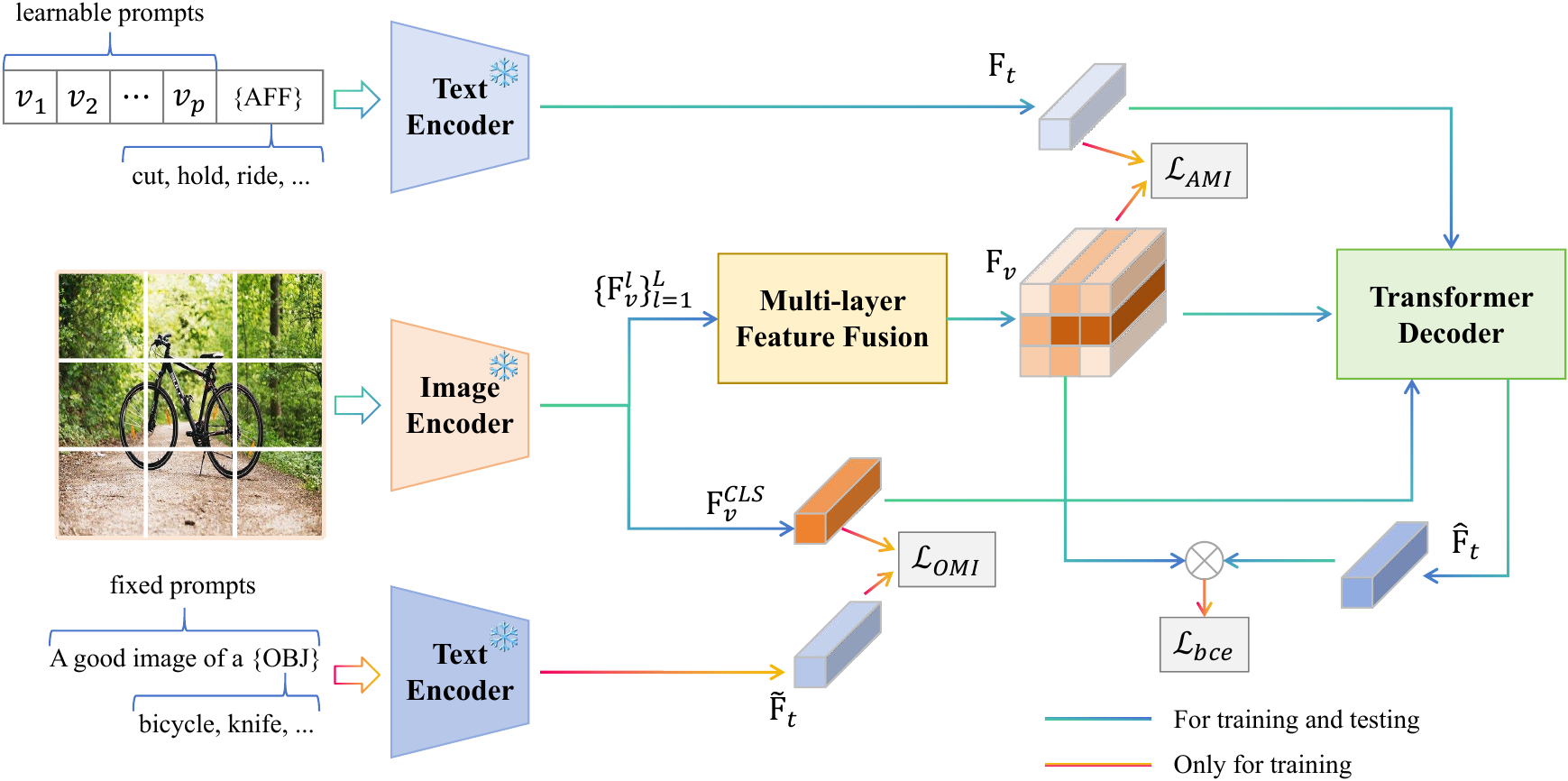}
  \caption{Overview of our proposed framework for text-guided affordance learning. During training, it involves two informative constraints, i.e. $\mathcal{L}_{AMI}$ and $\mathcal{L}_{OMI}$, to achieve text-image feature alignment in both affordance and object levels.}
  \label{fig:framework} 
\end{figure*}

\subsection{Mutual Information Maximization}

Mutual information (MI), a fundamental concept in information theory that measures the statistical dependence between random variables \cite{Cover2006}, has emerged as a powerful principle for representation learning. The InfoMax principle \cite{linsker1988self} posits that learning good representations involves maximizing MI between inputs and their representations. In deep learning, estimating and maximizing MI for high-dimensional data is challenging. Deep 
Contrastive Predictive Coding (CPC) \cite{oord2018representation}, though originally applied to sequences, inspired image-based approaches by maximizing MI between a ``context" (e.g., a patch) and its ``future" (e.g., a neighboring patch) using a contrastive loss, which provides a lower bound on MI. These works established the foundation and practical frameworks for effectively maximizing MI in representation learning, which our method builds upon.

Extending MI maximization to align different modalities has shown significant promise. 
In cross-modal tasks, maximizing MI between paired modalities serves as a critical optimization objective to align their latent spaces. For instance, in image-text retrieval, methods like CLIP \cite{radford2021learning} implicitly maximize MI by contrastive learning, pulling positive image-text pairs closer while pushing negatives apart in a shared embedding space. Similarly, Zhang et al. \cite{zhang2021cross} handle text-to-image generation task by explicitly maximizing the lower bound of MI between image and text using InfoNCE loss.

Inspired by the success of MI maximization in cross-modal feature alignment, we design MI-based constraints to pull the visual features of affordance regions and the corresponding text features closer, thereby boosting the performance of text-guided affordance learning.

\section{Methods} 

\subsection{Task Formulation}
Given an input image $\mathcal{I} \in \mathbb{R}^{H \times W \times 3}$ that contains objects with multiple affordance regions, our goal is to predict pixel-level affordance labels through continuous-valued masks $\{\mathbf{M}_c\}_{c=1}^C$ where $\mathbf{M}_c \in [0, 255]^{H \times W}$ denotes the grayscale intensity at each pixel location. Note that $0$ means non-affordance area and $(0, 255)$ denotes the affordance region intensity for category $c$. 

\subsection{Model Architecture}
Following Li et al. \cite{li2024one}, we focus on one-shot affordance learning with foundation models. As shown in Fig. \ref{fig:framework}, the overall framework is composed of four core components:

\begin{itemize}
    \item \textbf{Text Encoder}: In this paper, we have two text encoders to extract text features for affordance identification and object classification. The two text encoders are both instantiated as the textual branch of the CLIP model, which takes textual prompts as input to output robust text features. To adapt the CLIP text encoder to affordance learning, the prompt for the $i$-th affordance category is defined as the concatenation of a set of learnable context vectors $v = \{v_1, v_2, ..., v_p\}$ and the affordance-level class token embedding $c_i$. For the object branch, we use a fixed prompt template ``A good image of a [OBJ]" to construct the prompts for each object category. Here, the text features for affordance categories and object categories are denoted as $\mathbf{F}_t$ and $\tilde{\mathbf{F}}_t$ respectively. 
    \item \textbf{Image Encoder}: The backbone of DINOv2 is utilized to extract patch-level features for image input. For an input image $\mathcal{I} \in \mathbb{R}^{H \times W \times 3}$, the image encoder is able to output multi-layer feature maps $\{\mathbf{F}_v^{l}\}_{l=1}^L$, where $\mathbf{F}_v^{l} \in \mathbb{R}^{\hat{H}\hat{W}*D_v}$, and $\hat{H}$, $\hat{W}$, and $D_v$ indicate the height, width, and dimension of the feature map, and $L$ denotes the number of features used in the following multi-layer feature fusion. Additionally, the visual [CLS] token $\mathbf{F}_v^{CLS} \in \mathbb{R}^D_v$, a feature embedding to represent the semantic class of the object, can also be obtained by the image encoder.
    \item \textbf{Multi-layer Feature Fusion Module}: To process the multi-layer features generated by the image encoder, a trainable network is designed to merge the features from different layers. The multi-layer feature fusion module takes the multi-layer feature maps $\{\mathbf{F}_v^{l}\}_{l=1}^L$ as input, and outputs a fused visual feature map, which is then mapped to a feature dimension consistent with the text features. The final patch-wise visual features is denoted as $\mathbf{F}_v$.
    \item \textbf{Transformer Decoder}: In \cite{li2024one}, a CLS-guided transformer decoder is designed to promote the cross-modal interaction between textual and visual branches. The transformer decoder has three inputs, i.e. text features $\mathbf{F}_t$, visual patch features $\mathbf{F}_v$, and the visual [CLS] token $\mathbf{F}_v^{CLS}$. It employs the cross-attention mechanism to obtain class-aware text features $\hat{\mathbf{F}}_t$. 
\end{itemize}

With the class-aware text features $\hat{\mathbf{F}}_t$ and the patch-wise visual feature $\mathbf{F}_v$, the final prediction can be computed by:
\begin{equation}
    \mathbf{\hat{M}}_c = \text{UpSample}(\sigma(\mathbf{F}_v {\hat{\mathbf{F}}_t}^{c^{\top}})),
\end{equation}
where $\sigma$ is the sigmoid activation function and  $\hat{\mathbf{F}}_t^c$ is the text features for the $c$-th affordance category. Then, to ensure that the predicted affordance mask $\mathbf{\hat{M}}_c$ is consistent with the ground-truth mask $\mathbf{M}_c$, the trainable components of the model, i.e. the multi-layer feature fusion module and the transformer decoder, are optimized using a binary cross-entropy (BCE) loss defined as:
\begin{equation}
    \mathcal{L}_{\text{bce}} = -\frac{1}{C} \sum_{c=1}^C \frac{1}{N} \sum_{i=1}^N \left[ y_i \log(\hat{p}_i) + (1-y_i) \log(1-\hat{p}_i) \right],
\end{equation}
where $y_i \in [0, 1]$ is the scaled ground-truth label at pixel $i$, $\hat{p}_i = \mathbf{\hat{M}}_c(i) \in [0,1]$ is the predicted probability, and $N = H \times W$ is the total number of pixels.

\subsection{MI-based Text-Image Alignment in Affordance Level}
Although the [CLS]-guided transformer decoder proposed in \cite{li2024one} can provide the training process with certain semantic priors by embedding the [CLS] token into the cross-attention mechanism, it does not explicitly constrain feature-level alignment between text and image, which is essential for predicting the affordance mask in the text-guided affordance learning framework. Inspired by the applications of MI theory in representation learning and cross-modal feature alignment, we propose to introduce informative constraints to achieve text-image feature alignment in text-guided affordance learning. 

Considering that affordance regions are typically within and smaller than the object region, it is unreasonable to align the textual features corresponding to a certain affordance type with the visual features of the entire object region. Therefore, we propose to maximize the MI between text features and visual features of the smaller affordance region rather than the whole image. Formally, objective function can be expressed as:
\begin{equation}
\label{eq_ami}
    \mathcal{L}_{AMI} = I(\hat{\mathbf{F}}_t^c,\mathbf{F}_v^{aff})
\end{equation}
where $I(\cdot,\cdot)$ represents MI. $\mathbf{F}_v^{aff}$ and $\hat{\mathbf{F}}_t^c$ denote the visual features of the affordance areas and the text features of the corresponding affordance category respectively. Specifically, with the global patch-level visual features of the given image $\mathbf{F}_v$ and the ground-truth affordance mask $\mathbf{M}_c$, the visual features of the affordance areas can be obtained as follows:

\begin{equation}
\label{eq_maxpool}
    \tilde{\mathbf{M}}_c  = \text{MaxPool}(\mathbf{M}_c),
\end{equation}
\begin{equation}
    \mathbf{M}_c^{aff}  = \mathbf{1}_{\{\tilde{\mathbf{M}}_c > 0\}}, 
\end{equation}
\begin{equation}
\mathbf{F}_v^{aff} = \frac{
    \sum\limits_{i=1}^{\hat{H}} \sum\limits_{j=1}^{\hat{W}} 
    \mathbf{F_v(i, j)} \cdot \mathbf{1}[\mathbf{M}_c^{aff}(i,j)]
}{
    \sum\limits_{i=1}^{\hat{H}} \sum\limits_{j=1}^{\hat{W}} \mathbf{1}[\mathbf{M}_c^{aff}(i,j)]
}.
\end{equation}

Note that the usage of the MaxPool operation in Equ. (\ref{eq_maxpool}) is to resize the ground-truth map to match the size of the patch-level feature map $\mathbf{F}_v$, which enables spatial feature selection for each affordance instance.

Although maximizing MI between the two feature distributions is an effective approach to align them, directly estimating MI is challenging in high-dimensional settings. 
Hence, researchers prefer to maximize the lower bounds of MI rather than to perform a direct computation. In this paper, we choose InfoNCE \cite{oord2018representation}, one of the most common lower bounds of MI, to avoid directly computing $I(\hat{\mathbf{F}_t^c},\mathbf{F}_v^{aff})$. Then, the loss defined in Eq. (\ref{eq_ami}) is transferred to:
\begin{equation}
\mathcal{L}_{AMI} \approx 
  -\log \frac{
    \exp \left( s(\hat{\mathbf{F}}_t^c, \mathbf{F}_v^{aff}) / \tau_1 \right)
  }{
    \sum\limits_{k=1}^C \exp \left( s(\hat{\mathbf{F}}_t^{k}, \mathbf{F}_v^{aff}) / \tau_1 \right)
  },
\label{eq_ami_cl}
\end{equation}
where $s(\cdot,\cdot)$ means the similarity between two features, and $\tau_1$ is the temperature factor.
Through minimizing the $\mathcal{L}_{AMI}$ defined above, the model strives to achieve affordance-level feature alignment between text and image modalities during training, which will further boost the performance of text-guided affordance learning.

\subsection{MI-based Text-Image Alignment in Object Level}

In the OOAL framework \cite{li2024one}, the visual [CLS] token, i.e. $F_v^{CLS}$, is embedded as a semantic prior of the pre-trained model into the cross-attention mechanism. However, it overlooks the risk that, as training progresses, the cross-modal alignment at the object level may be disrupted, which could prevent the visual [CLS] token from providing effective guidance for affordance learning.
To address this limitation, we propose to ensure text-image alignment in object level by maximizing the MI between the visual [CLS] token and the textual prototype for the corresponding object category. Specifically, we begin by constructing a set of object-level text prompts based on the fixed template ``a good image of a [OBJ]", where [OBJ] represents the class names for a given object. Then, the text features for each object category can be extracted by using the text encoder. With the text features denoted as $\Tilde{\mathbf{F}}_t$, the objective function for text-image alignment in object level is expressed as:
\begin{equation}
    \mathcal{L}_{OMI} = I(\tilde{\mathbf{F}}_t^c,\mathbf{F}_v^{CLS}).
\end{equation}

Similar to the approximation for $\mathcal{L}_{AMI}$, we utilize InfoNCE to estimate the MI between the visual [CLS] token and the corresponding object-level textual prototype. Thus the text-image alignment loss in object level can be calculated by:

\begin{equation}
        \mathcal{L}_{OMI} \approx -\log \frac{
    \exp \left( s(\tilde{\mathbf{F}}_t^c, \mathbf{F}_v^{CLS}) / \tau_2 \right)
  }{
    \sum\limits_{k=1}^{C_{obj}}, \exp \left( s(\tilde{\mathbf{F}}_t^{k}, \mathbf{F}_v^{CLS}) / \tau_2 \right)
  },
\label{eq_omi_cl}
\end{equation}
where $C_{obj}$ represents the number of object classes and $\tau_2$ denotes the temperature factor.


With the MI-based constraint for text-image alignment in object level, the visual encoder is encouraged to output visual [CLS] tokens that align well with the object-level class prototypes. This can provide stronger semantic prior for the cross-attention in transformer decoder and facilitate the identification of affordance regions with textual guidance.

\subsection{Joint Training of the Whole Framework}
In previous sections, we have described in detail how we leverage mutual information to achieve cross-modal feature alignment. In conclusion, the overall training objective of the proposed framework includes three loss items:
\begin{equation}
\label{eq_total}
    \mathcal{L} = \mathcal{L}_{bce} + \lambda_1 \mathcal{L}_{AMI} + \lambda_2 \mathcal{L}_{OMI},
\end{equation}
where $\lambda_1$ and $\lambda_2$ are the weighting factors for balancing the three loss items. The values of $\lambda_1$ and $\lambda_2$ can be adjusted to obtain better performance.

\section{Experiments}

In this section, we conduct comparative experiments, ablation studies and parameter sensitivity analysis to demonstrate the effectiveness of the proposed method. The experimental settings, results and corresponding analysis are provided below.

\subsection{Experimental Settings}

    \subsubsection{Dataset} We utilize the widely used affordance grouding dataset AGD20K for evaluation. AGD20K consists of over 20k images of 36 affordances and 50 object categories, and the images are further split into Seen and Unseen for different settings. As the original training set of AGD20K has only image-level annotations, Li et al. \cite{li2024one} manually select and annotate 50 egocentric images for affordance learning in one-shot setting. In this paper, we use the one-shot dataset and follows \cite{li2024one} in all data-related settings.  
    \subsubsection{Implementation details} In the proposed method, the text encoders are the textual branch of pre-trained CLIP model, while the image encoder is from a pre-trained DINOv2 model with a base vision transformer (ViT-B). The parameters of both the text and image encoders are frozen during training. With the pre-trained models, we train the remaining modules for 20k iterations in one-shot manner. The SGD optimizer with a learning rate of 0.01 is adopted for training. For quantitative evaluation, we compute three widely used metric scores based on the prediction results, i.e. Kullback-Leibler Divergence (KLD), Similarity (SIM), and Normalized Scanpath Saliency (NSS). Lower KLD scores, along with higher SIM and NSS scores, indicate better performance. In the testing stage, we load the best checkpoint saved during training and test the trained model on AGD20K testing set, following what OOAL \cite{li2024one} done. 

\subsection{Comparisons}

We compare the proposed framework with four state-of-the-art methods for affordance learning, including MaskCLIP \cite{dong2023maskclip}, SAN \cite{xu2023side}, ZegCLIP \cite{zhou2023zegclip} and OOAL \cite{li2024one}. The comparative experiments were conducted in one-shot setting, where only a single sample for each object class is used for training. The comparison results on AGD20K dataset are shown in Tables \ref{comparison1} and \ref{comparison2}. 

\begin{table}[h]
\caption{Comparison with state-of-the-art methods on one-shot Seen split of AGD20K dataset.}
\label{comparison1}
\begin{center}
\begin{tabular}{|c||c|c|c|}
\hline
Method & KLD $\downarrow$ & SIM $\uparrow$ & NSS $\uparrow$ \\
\hline
MaskCLIP \cite{dong2023maskclip} & 5.752 & 0.169 & 0.041 \\
SAN \cite{xu2023side} & 1.435 & 0.357 & 0.941 \\
ZegCLIP \cite{zhou2023zegclip} & 1.413 & 0.387 & 1.001 \\
OOAL \cite{li2024one} & 0.740 & 0.577 & 1.745 \\
Ours & \textbf{0.719} & \textbf{0.581} & \textbf{1.787}  \\
\hline
\end{tabular}
\end{center}
\end{table}

\begin{table}[h]
\caption{Comparison with state-of-the-art methods on one-shot Unseen split of AGD20K dataset.}
\label{comparison2}
\begin{center}
\begin{tabular}{|c||c|c|c|}
\hline
Method & KLD $\downarrow$ & SIM $\uparrow$ & NSS $\uparrow$ \\
\hline
MaskCLIP \cite{dong2023maskclip} & 6.052 & 0.152 & 0.047 \\
SAN \cite{xu2023side} & 1.580 & 0.351 & 1.022 \\
ZegCLIP \cite{zhou2023zegclip} & 1.552 & 0.361 & 1.042 \\
OOAL \cite{li2024one} & 1.070 & 0.461 & 1.503 \\
Ours & \textbf{1.023} & \textbf{0.482} & \textbf{1.535}  \\
\hline
\end{tabular}
\end{center}
\end{table}

Obviously, the proposed method outperforms all comparative methods in all three metrics across different data splits, indicating state-of-the-art performance in one-shot affordance learning with textual guidance. In detail, as SAN utilizes only the visual prior of the CLIP model, rather than the multi-modal prior, it cannot demonstrate satisfactory capability in identifying affordance areas. In contrast, MaskCLIP and ZegCLIP both leverage the multi-modal prior of the foundation model, but fail to effectively adapt to the affordance learning task. Although OOAL produces the most similar results to our method among the comparative approaches, our method comprehensively outperforms OOAL. This is primarily because our method better preserves the text-image alignment property in feature level, which is highly beneficial for leveraging text guidance in visual affordance learning.

\subsection{Ablation Studies}
We further highlight the significance of our proposed informative constraints through ablation studies regarding the loss items in Eq. (\ref{eq_total}). The experimental results have been provided in Table \ref{ablation}.

By adding the affordance-level MI maximization loss $\mathcal{L}_{AMI}$ to the basic binary cross entropy loss $\mathcal{L}_{bce}$, the model achieves obvious better performance in both KLD and NSS, with comparable result in SIM. When adding only the loss of MI maximization in object level, i.e. $\mathcal{L}_{OMI}$, the SIM score is improved and better overall performance is achieved. The best performance occurs when all three loss items are used to optimize the model, indicating that the two MI-based constraints proposed in this paper boost text-guided affordance learning in a complementary manner.

\begin{table}[h]
\caption{Ablation studies on one-shot Seen split of AGD20K dataset.}
\label{ablation}
\begin{center}
\begin{tabular}{|l||c|c|c|}
\hline
Method & KLD $\downarrow$ & SIM $\uparrow$ & NSS $\uparrow$ \\
\hline
$\mathcal{L}_{bce}$ & 0.740 & 0.577 & 1.745 \\
+ $\mathcal{L}_{AMI}$ & 0.730 & 0.575 & 1.772 \\
+ $\mathcal{L}_{OMI}$ & 0.735 & \textbf{0.581} & 1.755 \\
+ $\mathcal{L}_{AMI}$ + $\mathcal{L}_{OMI}$ &  \textbf{0.719} & \textbf{0.581} & \textbf{1.787} \\

\hline
\end{tabular}
\end{center}
\end{table}

\begin{figure*}[htbp]
  \centering            
  \subfloat[]{
  \label{fig:tau_KLD}
  \includegraphics[width=0.32\textwidth]{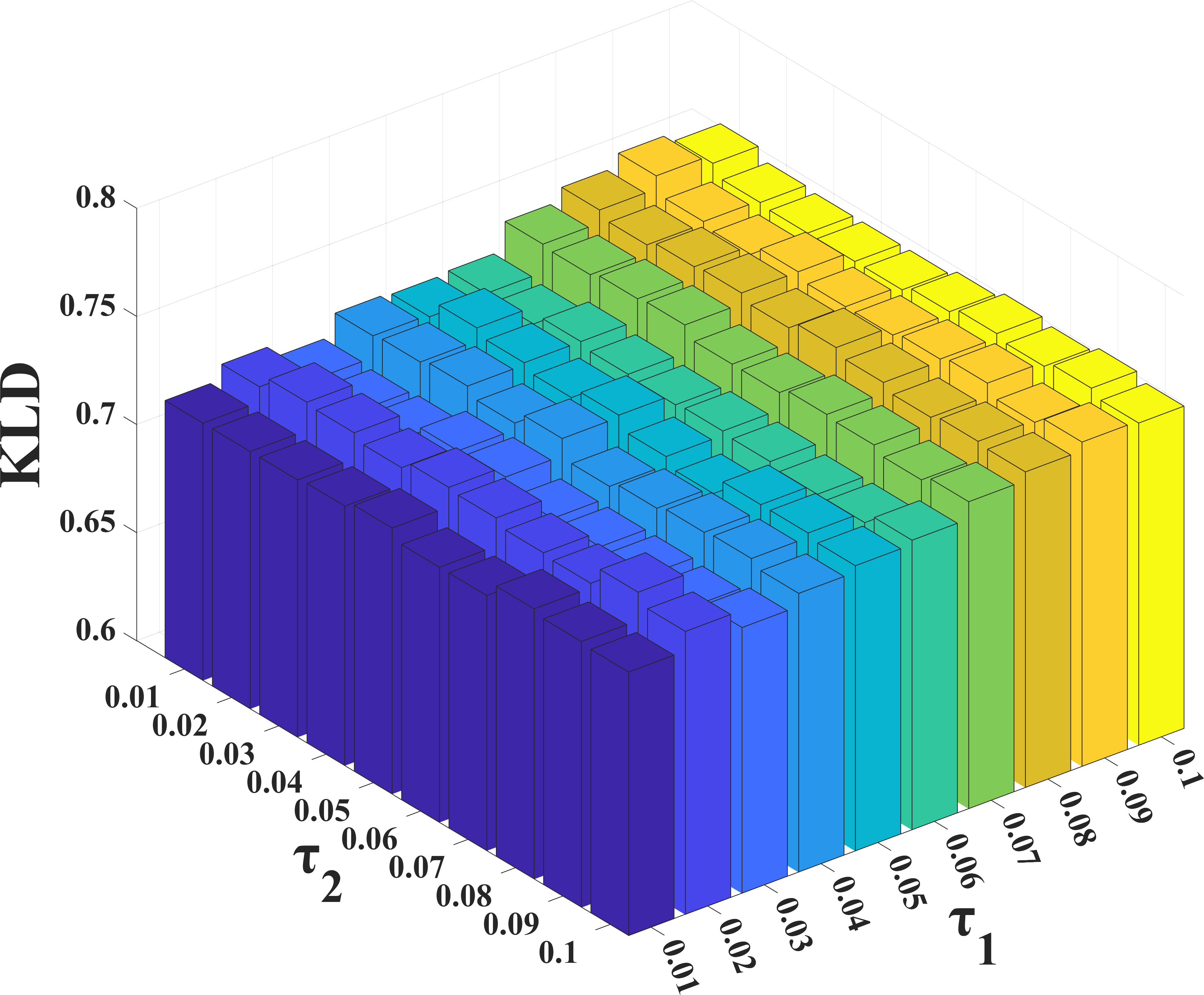}
}\hspace{-2mm}
  \subfloat[]{
  \label{fig:tau_SIM}
  \includegraphics[width=0.32\textwidth]{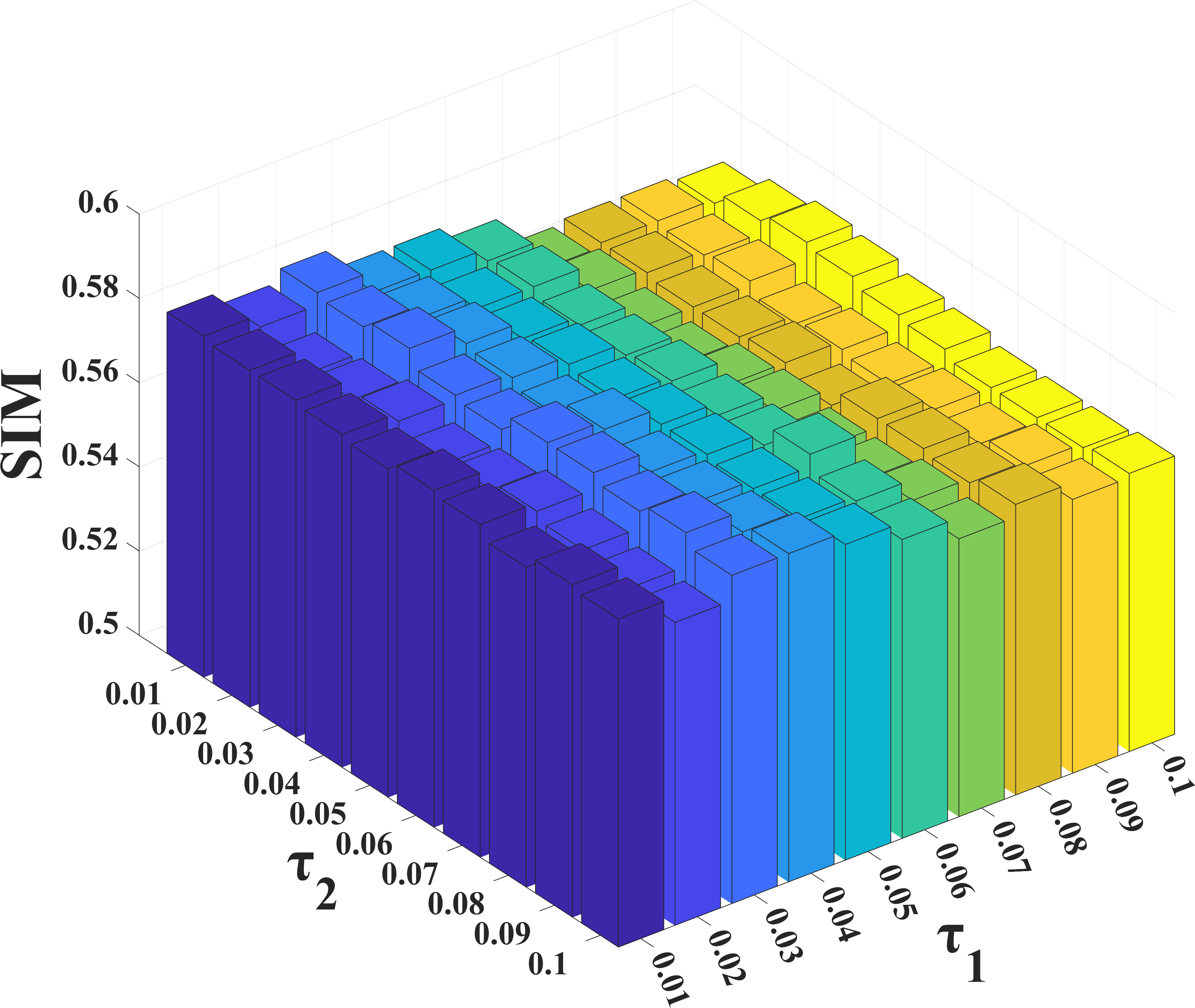}
}\hspace{-3mm}
  \subfloat[]{
  \label{fig:tau_NSS}
  \includegraphics[width=0.32\textwidth]{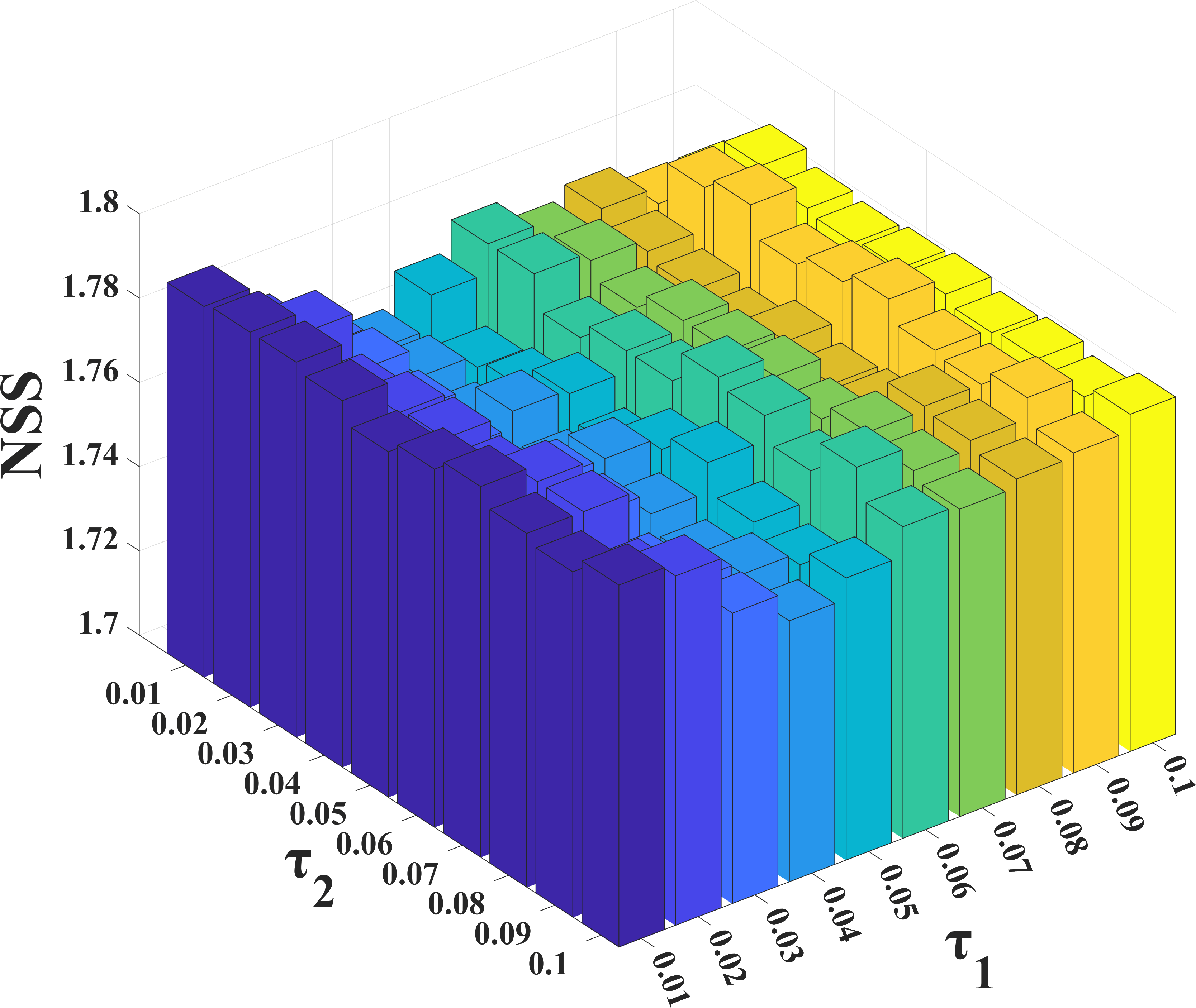}
}
  \caption{The performance of the proposed method when $\tau_1$ and $\tau_2$ are varied.}
  \label{fig:tau} 
\end{figure*}

\begin{figure*}[htbp]
  \centering            
  \subfloat[]{
  \label{fig:lambda_KLD}
  \includegraphics[width=0.32\textwidth]{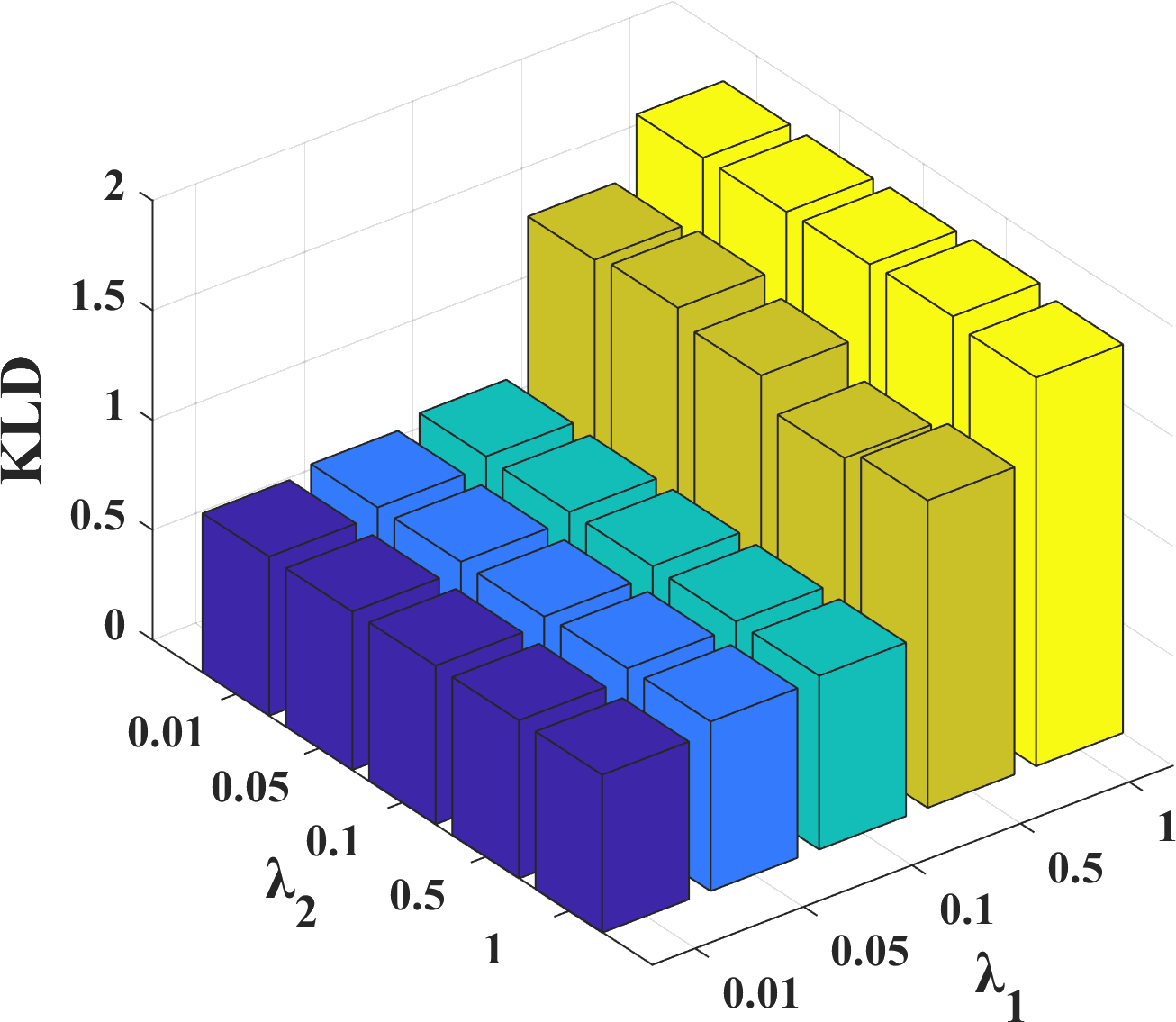}
}\hspace{-2mm}
  \subfloat[]{
  \label{fig:lambda_SIM}
  \includegraphics[width=0.32\textwidth]{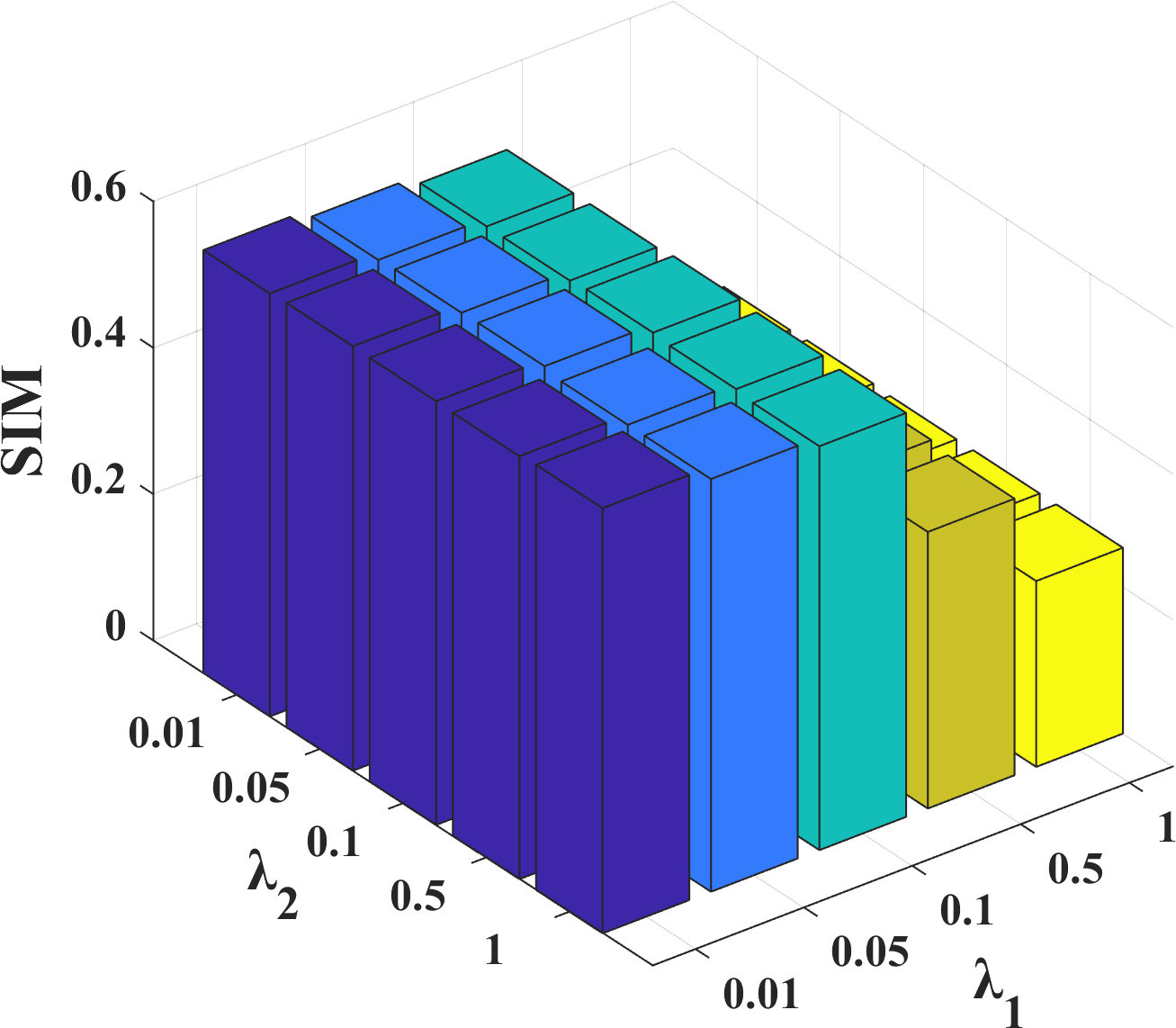}
}\hspace{-3mm}
  \subfloat[]{
  \label{fig:lambda_NSS}
  \includegraphics[width=0.32\textwidth]{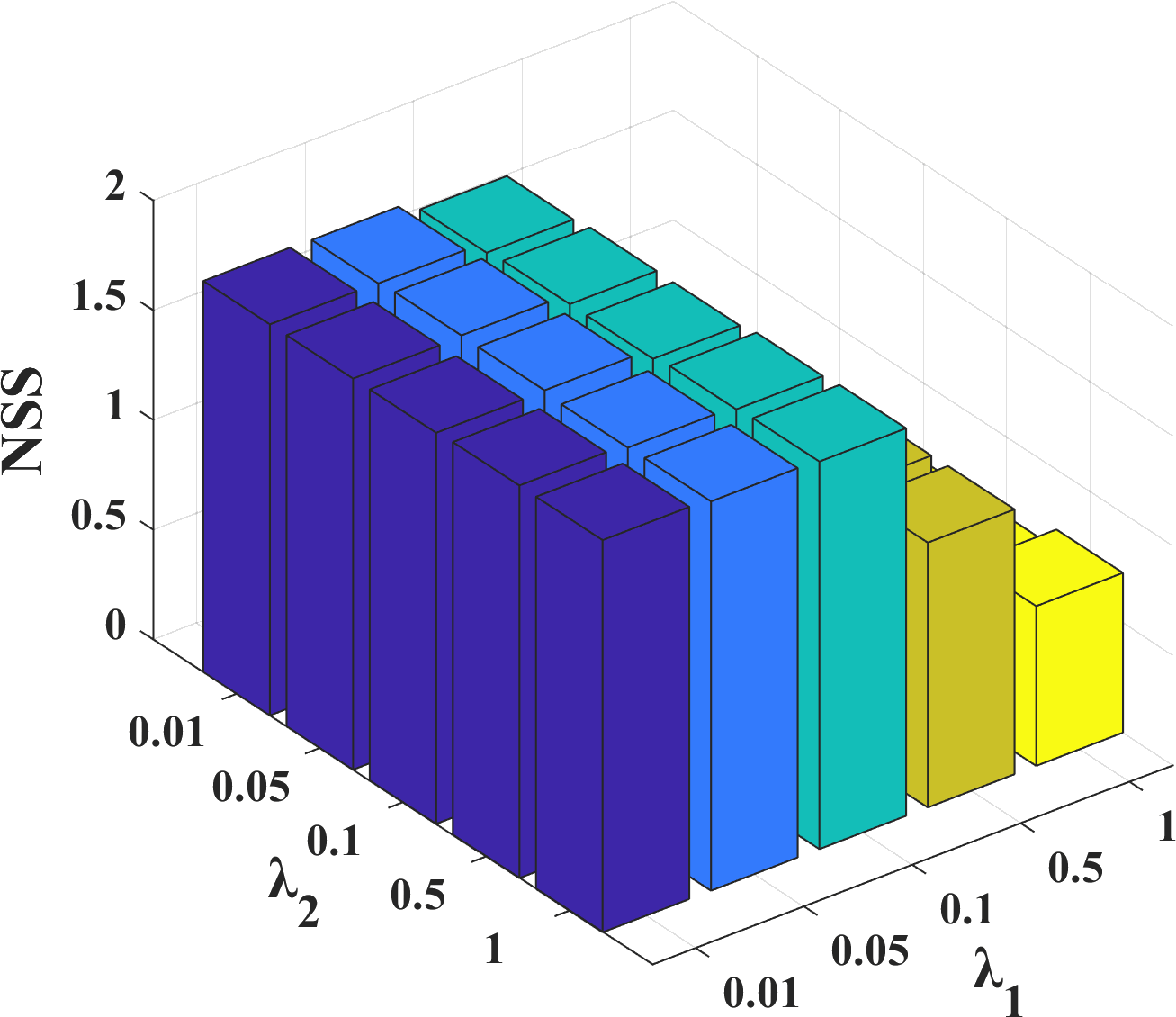}
}
  \caption{The performance of the proposed method when $\lambda_1$ and $\lambda_2$ are varied.}
  \label{fig:lambda} 
\end{figure*}

\subsection{Parameter Sensitivity Analysis}
\subsubsection{Effect of Temperature Factors}
The temperature parameters, i.e. $\tau_1$ and $\tau_2$ in Equ. (\ref{eq_ami_cl}) and Equ. (\ref{eq_omi_cl}), are used to modulate the gradients of samples during training. To explore the effect of the two temperature parameters on our proposed method, we employ a grid search strategy in the range of $[0.01, 0.1]$ to find the appropriate settings of $\tau_1$ and $\tau_2$. The experimental results are shown in Fig. \ref{fig:tau}. It is obvious that the overall performance is much better when $\tau_1$ is set to a smaller value, with the best performance observed at 0.01. Although our method's performance in affordance identification is less sensitive to the variation of $\tau_2$ compared to $\tau_1$, the results still consistently favor a smaller value for it. This is mainly because small temperature factors amplify the gradients from hard negatives, forcing the model to learn more fine-grained discriminative features by aggressively separating similar but distinct affordance-level or object-level instances. 
Therefore, we set both temperature parameters $\tau_1$ and $\tau_2$ to 0.01 in our method, as this configuration can yield the optimal performance in identifying affordance areas in an image.

\subsubsection{Effect of Weighting Factors}
As shown in Equ. (\ref{eq_total}), there are two weighting parameters, i.e. $\lambda_1$ and $\lambda_2$, in the proposed framework to balance different loss items during the joint training stage. To figure out the effect of the weighting parameters on the performance of affordance identification, we implement our method with different combinations of $\lambda_1$ and $\lambda_2$, the results of which have been presented in Fig. \ref{fig:lambda}. Specifically, we tune $\lambda_1$ and $\lambda_2$ over a grid of values: $[0.01, 0.05, 0.1, 0.5, 1.0]$ to explore the best combination of weighting factors. From the subfigures we can see that the performance of the proposed method exhibits a strong dependence on $\lambda_1$, while remaining relatively robust to variations in $\lambda_2$ across the tested range. This phenomenon is primarily due to the fact that the value of $\mathcal{L}_{OMI}$ consistently remains at a low magnitude during training compared to the other two losses. However, this does not justify removing this constraint, as its necessity has been confirmed through ablation studies. Based on the experimental results shown in Fig. \ref{fig:lambda}, we set $\lambda_1=0.01$ and $\lambda_2=1.0$ to enhance the performance of affordance identification.

\subsection{Visualization Results}

\begin{figure}[htbp]
  \centering            
  \includegraphics[width=0.48\textwidth]{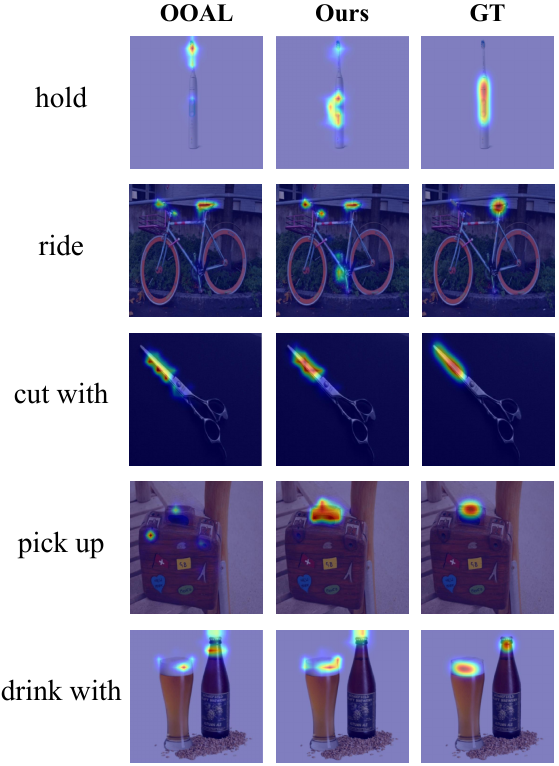}
  \caption{Qualitative results of our method and OOAL \cite{li2024one} on AGD20K dataset. Our proposed method predicts more similar affordance masks with ground truths (GT).}
  \label{fig_vis} 
\end{figure}

In Fig. \ref{fig_vis}, we showcase the qualitative results produced by our proposed method along with those of OOAL \cite{li2024one}. Compared to OOAL, our method is able to produce affordance masks that are more similar to the ground truths. For example, Our method accurately identifies the area for ``hold'' in the electric toothbrush and the area for ``pick up" in the suitcase, while OOAL fails in these cases. Additionally, as shown in the second row of the figure, our method effectively highlights the bicycle's handlebars, saddle, and pedals for the ``ride" affordance, which is more reasonable even though there are some differences from the ground truth. In summary, the visual results further validate the effectiveness of the proposed method and demonstrate its superiority over existing approaches.

\section{CONCLUSIONS}

In this paper, we propose a novel informative affordance learning method that ensures text-image alignment to facilitate the identification of affordance areas with textual guidance. In specific, mutual information maximization techniques are employed to achieve cross-modal feature alignment in both affordance level and object level. This method is effective in one-shot affordance learning with textual guidance as it preserves and utilizes the multi-modal priors in foundation models. In the future, we will extend the application scenarios of our method and validate its effectiveness in real-world robotic tasks.

\addtolength{\textheight}{-12cm}   




\bibliographystyle{IEEEtran}
\bibliography{IEEEtranBST/IEEEabrv, IEEEtranBST/IEEEexample}

\end{document}